\documentclass{article}

\usepackage{PRIMEarxiv}
\usepackage{placeins}  
\usepackage[utf8]{inputenc} 
\usepackage{float} 
\usepackage[T1]{fontenc}    
\usepackage{hyperref}       
\usepackage{url}            
\usepackage{booktabs}       
\usepackage{amsfonts}       
\usepackage{nicefrac}       
\usepackage{microtype}      
\usepackage{lipsum}
\usepackage{fancyhdr}       
\usepackage{graphicx}       
\graphicspath{{media/}}     
\usepackage{multirow}
\usepackage{gensymb}
\usepackage{amsmath}       
\usepackage{algorithm}     
\usepackage{algpseudocode} 
\usepackage{float}  
\usepackage{threeparttable}
\usepackage{caption}
\usepackage{tcolorbox}
\usepackage{authblk}
\usepackage{comment}           
\usepackage{natbib}
\usepackage{amsmath}

\title{Neuroscience Inspired Graph Operators Towards Edge-Deployable Virtual Sensing for Irregular Geometries}

\author[1]{William Howes}
\author[1]{Farid Ahmed}
\author[1]{Kazuma Kobayashi}
\author[1,3,4]{Souvik Chakraborty}
\author[1,2]{Syed Bahauddin Alam\textsuperscript{*}}

\affil[1]{Grainger College of Engineering, Nuclear, Plasma \& Radiological Engineering Department, University of Illinois Urbana-Champaign, Urbana, IL, USA
}

\affil[2]{National Center for Supercomputing Applications, Urbana, IL, USA
}

\affil[3]{Department of Applied Mechanics, Indian Institute of Technology Delhi, New Delhi, India
}

\affil[4]{Yardi School of Artificial Intelligence, Indian Institute of Technology Delhi}

\affil[*]{Corresponding author: \href{mailto:alams@illinois.edu}{alams@illinois.edu}}

\begin{document}
\maketitle
\begin{abstract}
Predicting full-field physics through the real-time virtual sensing of engineering systems can enhance limited physical sensors but often requires sparse-to-dense reconstruction, complex multiphysics, and highly irregular geometries as well as strict latency and energy constraints for edge-deployability. Neural operators have been presented as a potential candidate for such applications but few architectures exist that explicitly address power consumption. Spiking neuron integration can provide a potential solution when integrated on neuromorphic hardware but the current existing neuron models result in severe performance degradation towards regression-based virtual sensing. To address the performance concerns and edge-constraints, we present the Variable Spiking Graph Neural Operator (VS-GNO) which integrates a sophisticated spectral-spatial convolutional analysis and a previously developed Variable Spiking Neuron (VSN) and energy-error balance loss function. With a non-spiking $L_2$ error baseline of $0.4\%$, VS-GNO can provide a reconstruction error of $0.71\%$ with $15\%$ average spiking in its spectral-only form and $1.04\%$ with $24.5\%$ spiking in its entire form. These results position VS-GNO as a promising step towards energy-efficient, edge-deployable neural operators for real-time sparse-to-dense virtual sensing in complex, highly irregular engineering environments.
\end{abstract}

\vspace{-1mm}
\section{Introduction}


To avoid frequent interference of normal operations for engineering systems such as nuclear reactors, understanding and predicting full-field physics that is unmeasurable by limited physical instruments, such as coolant flow or turbulence, is vital and can provide insight into material degradation and remaining useful life \cite{kobayashi2024virtualsensingenablerealtime}. This framework encapsulates the definition and motive behind virtual sensing. Traditional methodologies and solvers can provide high-fidelity reconstruction of the governing physics but do not satisfy the strict edge deployment constraints of various engineering applications which require low latency and low power for real-time, efficient monitoring \cite{kobayashi2024virtualsensingenablerealtime, howes2026graphneuraloperatoredge}. 
To address these challenges, sophisticated neural operator architectures, such as the Fourier Neural Operator (FNO) \cite{DBLP:journals/corr/abs-2010-08895}, provide faster reconstruction performance for full-field physics by approximating the operator solution for the governing equations through data-driven backpropagation but struggle to handle highly irregular geometries and complex multiphysics with limited cross-domain input \cite{howes2026graphneuraloperatoredge, kobayashi2024virtualsensingenablerealtime}. As a result, the development of neural operator architectures for feasible virtual sensing within engineering applications must satisfy the following: sparse-to-dense full-field reconstruction (sparse input to dense field output) for complex multiphysics on irregular geometries that is edge deployable and fulfills strict computational requirements.



Spiking neurons are a promising solution to the reduced energy requirements of neural operator virtual sensing \cite{brainsci12070863}. Spiking Neural Networks (SNNs) which utilize spiking neurons rely on biologically-inspired, event-driven, asynchronous communication where neurons communicate only when a threshold is reached, providing significantly reduced energy consumption with neuromorphic hardware \cite{brainsci12070863}. The most popular spiking neuron model is the Leaky Integrate and Fire (LIF) but utilizing a fully-trained LIF model for regression-based neural operators has been shown to have severe performance degradation and significantly more when the input is rate encoded \cite{garg2023neuroscienceinspiredscientificmachinepart2}. To address the shortcomings of LIF, a Variable Spiking Neuron (VSN) was presented and integrated into a Wavelet Neural Operator (VS-WNO) \cite{garg2023neuroscienceinspiredscientificmachinepart2}. The model was trained from scratch with surrogate gradients using an energy-balance loss function for optimizing model weights as well as threshold and leakage. VS-WNO provided improved field reconstruction performance with the potential for energy reduction through variable spiking that naturally handles direct encoding and results in minimal degradation for regression-based tasks at low spiking time steps. Use cases for the VS-WNO were limited to structured grids with dense-to-dense operator learning. 


In this paper, we present the Variable Spiking Graph Neural Operator (VS-GNO) which, to our knowledge, is the first among graph networks applied to operator-based virtual sensing. This model directly addresses sparse-to-dense, multi-output virtual sensing with irregular geometries and is evaluated with a use case that aligns with those characteristics and goes beyond typical modeling benchmarks for spiking neural operators. The VS-GNO utilizes the architecture of the Virtual Irregular Real-Time Sparse Operator (VIRSO) \cite{howes2026graphneuraloperatoredge} and combines it with the spiking behavior of the VSN and the energy-balance loss function for optimizing trainable weight, threshold, and leakage parameters. In summary, this work provides the following contributions: a unique spiking graph operator towards natural, energy-efficient handling of irregular geometries and a repositioning of variable spiking and its corresponding loss function towards edge-deployable, real-time virtual sensing which consists of sparse inputs and complex physics on unstructured grids with strict computational constraints.

\section{Methods}

\subsection{Problem Formulation}

\begin{figure}[h]
  \centering
  \includegraphics[width=\linewidth]{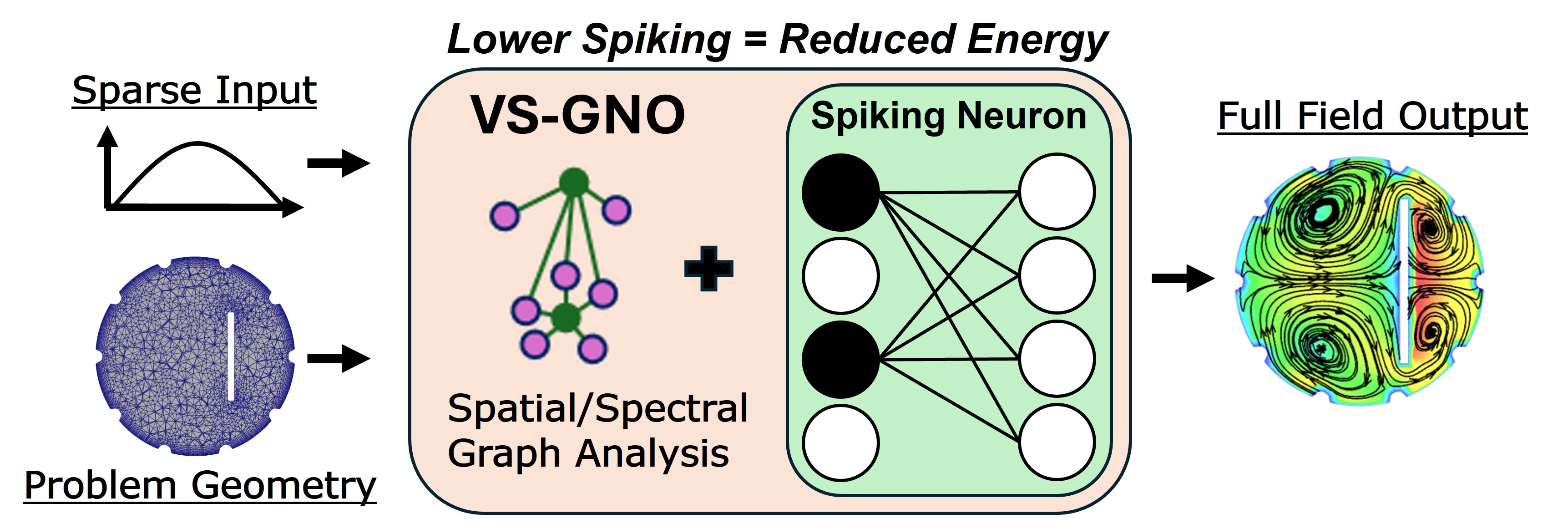}
  \caption{VS-GNO for Sparse-To-Dense Reconstruction on Irregular Grids with Reduced Energy Consumption}
  \label{fig:vsgno}
\end{figure}

The sparse-to-dense reconstruction of VS-GNO, illustrated in Figure \ref{fig:vsgno}, learns the following operator mapping $\mathcal{G}$:

\begin{equation}
\mathcal{G}: \mathcal{U} \to \mathcal{S}, \quad \mathcal{G}(\mathbf{u})(\mathbf{x}) = \mathbf{s}(\mathbf{x}), \quad \mathbf{x}\in\mathcal{Y}.
\label{eq:framework}
\end{equation}

$\mathbf{s}(\mathbf{x})\in\mathbb{R}^k$ represents the $k$ solution outputs to the governing equations, such as pressure or temperature, evaluated at $\mathbf{x}$ within our desired domain $\mathcal{Y}$. $\mathbf{u} = [u_1, \dots, u_b] \in \mathcal{U}$ represents $b$ multi-modal sparse cross-domain input spaces typically defined along the geometry boundary. $u_i$ can be a scalar $\mathbb{R}$, such as inlet velocity, or a functional space $L^2(D')$ defined typically on a separate domain than $\mathcal{Y}$. VS-GNO approximates Equation \ref{eq:framework} by approximating $L$ consecutive non-linear integral mappings defined by the following:

\begin{equation} \label{eq:kernel_int}
\mathbf{v}_{\ell+1}(\mathbf{x}) = \sigma(W\mathbf{v}_\ell(\mathbf{x}) + \int_{\mathcal{Y}}{\mathcal{K}_\phi}(\mathbf{x},\mathbf{z})\mathbf{v}_\ell(\mathbf{z})d\mathbf{z}),
\end{equation}

where $\mathbf{v}_\ell$ and $\mathbf{v}_{\ell+1}$ represent intermediate function evaluations on the desired domain $\mathbf{x} \in \mathcal{Y}$. $\sigma$ represents a nonlinear activation such as Sigmoid or ReLU that provides nonlinearity to our integral framework while $W \mathbf{v}_t(\mathbf{x})$ represents a residual connection and $\mathcal{K}_\phi$ represents the parameterized kernel utilized in the integral over the geometric domain $\mathcal{Y}$. 


\subsection{Variable Spiking Neuron}


To address the poor regression performance of the LIF neuron, the Variable Spiking Neuron (VSN) was introduced to provide a gradient variable spike output that is a continuous function of the original LIF spike and initial input, allowing for limited degradation in regression performance and better handling of direct encoding inputs for natively trained spiking networks. The VSN has similar temporal dynamics as the LIF neuron with leakage and memory accumulation over $T$ spike time steps (STS) but modifies the signal sent to forward neurons with the following framework \cite{garg2023neuroscienceinspiredscientificmachinepart2}:

\begin{equation}
\begin{gathered}
    M^{(t)} = \beta M^{(t-1)} + z^{(t)} \\
    \tilde{y}^{(t)} = 
    \left\{ 
        \begin{array}{lr} 
            1 ; \quad M^{(t)} \geq \Theta \\
            0 ; \quad M^{(t)} < \Theta 
        \end{array}
    \right\} \quad \text{if } \tilde{y}^{(t)}, M^{(t)} \gets 0\\
    y^{(t)} = \sigma(z^{(t)}\tilde{y}^{(t)}), \quad \text{where } \sigma(0) = 0, 
\end{gathered}    
\label{eq:vsn}
\end{equation}

where $\Theta$ represents a trainable threshold for a spiking event and $\beta$ is a trainable leakage parameter that controls the influence of past memory $M^{(t-1)}$. $y^{(t)}$ represents the final output communicated to forward neurons which is a continuous function $\sigma$ (linear or nonlinear) of the binary LIF spike $\tilde{y}^{(t)}$ weighted by the original input \cite{garg2023neuroscienceinspiredscientificmachinepart2}. The requirement that $\sigma(0)=0$ allows for no communication in the event that the threshold is not reached \cite{garg2023neuroscienceinspiredscientificmachinepart2}. To avoid precision loss, $z^{(t)}$ is implemented with direct encoding. As a result, the VSN is able to prevent significant regression loss degradation. Although, the energy consumption for VSN is larger than the original LIF structure by a factor of 1.7 (based on post-layout SpiNNaker2 analysis), the improvement in field reconstruction error significantly outweighs the minimal loss in efficiency \cite{garg2023neuroscienceinspiredscientificmachinepart2}.


\subsection{VS-GNO Algorithm}

To perform full-field reconstruction from sparse information, shown in Figure \ref{fig:vsgno}, the VS-GNO transforms the desired evaluation points to a graph $G=(\mathbb{V},\mathbb{E})$ with a KNN algorithm based on the inverse euclidean distance metric. Latent embeddings of the discretized input representation ($\mathbf{u}_q\in\mathbb{R}^q$) are calculated with a fully-connected network (FCN) projection $M$ and combined with the node coordinates to define the initial node features $\mathbb{X}$. An FCN-based mapping $P$ then lifts $\mathbb{X}$ to the starting intermediate state $\mathbf{v}_0(\mathcal{Y}_n)\in \mathbb{R}^{n\times d_v}$

VS-GNO then employs $L$ consecutive convolutional integral layers, defined by Equation \ref{eq:kernel_int} and approximated by a graph spectral-spatial collaboration. The spectral portion provides global information by taking the partial Graph Fourier Transform ($\mathbf{Q}_m^T$ for $m$ lowest frequency modes) of the input state, applying an enriched, directly-parameterized kernel $\mathbf{K}$, and finally returning to the spatial domain with the partial inverse Graph Fourier Transform ($\mathbf{Q}_m$). In addition, a linear residual skip $w(\mathbf{v_\ell})$ is added before a final nonlinear activation function $\sigma$. The spatial graph analysis, which provides missing high-frequency information, consists of applying a trainable weight matrix $\mathbf{W}$ and performing neighborhood feature aggregation for each node. The accumulation is modified by a trainable gating mechanism matrix $\Gamma$ where the contribution of each neighbor $v$ for node $u$ is weighted by $\Gamma_{uv}$, allowing for graph structure optimization during training. To define the next iterative state $\mathbf{v}_{\ell+1}$, the spatial and spectral outputs are concatenated and fed through a linear projection mapping $f$ and added to the previous layer $\mathbf{v}_\ell$ with a final residual skip. After $L$ layers, the final state $\mathbf{v}_L$ is downlifted, with an FCN $Q$, to the final multi-output: $\mathbf{s}(\mathcal{Y}_n)$.

VS-GNO applies the VSN in Equation \ref{eq:vsn} after various computational layers within the above graph architecture, replacing any previous activation functions $\sigma$ that might have existed. Let $\mathcal{V}(\mathbf{\Theta}, \vec{\beta})$ represent a VSN layer with threshold and leakage vectors of size equal to the input feature dimension of each node. Defining unique thresholds for each feature of a node allows for geometry generalization but can risk severe performance degradation with small dimensions and their lack of expressiveness in threshold and leakage configuration. With spiking, we deviate from the original final downlift $Q$ such that the final state $\mathbf{v}_L^{(t)}$ at each spike step is passed through the first linear layer of $Q$ and a VSN to produce $\tilde{\mathbf{s}}^{(t)}$, which is averaged over all $T$ steps and then fed through the final linear layer to obtain $\mathbf{s}(\mathcal{Y}_n)$. The following framework summarizes the VS-GNO algorithm:

\begin{equation}
\begin{gathered}
\quad P=\mathcal{V}_{P, 2}(W_2\mathcal{V}_{P, 1}(W_1(\dots))), \quad M=W_2\mathcal{V}_{M}(W_1(\dots)), \\
\mathbf{v}_0^{(t)}(\mathcal{Y}_n) = P(\mathbb{X}), \quad \mathbb{X}_i = \{\mathbf{x}_i, M(\mathbf{u_q})\}, \\
\mathbf{v}_{\ell+1}^{\text{spatial}, (t)}= \mathcal{V}_{\text{spatial},\ell,2}(\mathbf{\Gamma}  \odot \mathbf{A}\mathcal{V}_{\text{spatial},\ell,1}(\mathbf{v}_{\ell}^{(t)}\mathbf{W}))\ \\
\mathbf{v}_{\ell+1}^{\text{spectral}, (t)} = \mathcal{V}_{\text{spectral},\ell}(\mathbf{Q}_m \cdot\mathbf{K}\times_1\mathbf{Q}^T_m \cdot \mathbf{v}_{\ell}^{(t)} + w(\mathbf{v}_{\ell}^{(t)})) \\
\mathbf{v}_{\ell+1}^{(t)} = 
\mathcal{V}_{f,\ell}(f([\mathbf{v}_{\ell+1}^{\text{spatial}, (t)} ||  \mathbf{v}_{\ell+1}^{\text{spectral}, (t)}]) + \mathbf{v}_{\ell}^{(t)}) \\
\tilde{\mathbf{s}}^{(t)} = \mathcal{V}_\text{final}(W_1\mathbf{v}_L^{(t)}) \to 
\mathbf{s}(\mathcal{Y}_n) = W_2\frac{1}{T}\Sigma_{t=1}^T\tilde{\mathbf{s}}^{(t)}.
\end{gathered}
\label{eq:vsgno}
\end{equation}

$\mathbf{A}$ is the adjacency matrix, $\odot$ is the Hadamard product, $||$ represents concatenation, and $\times_1$ is a mode-1 tensor product. The subscripts for $\mathcal{V}$ define unique VSN layers within the architecture.

\subsection{Spiking Loss Term}

The following loss function is utilized from the VS-WNO for optimizing the trainable weight, threshold, and leakage parameter configuration \cite{garg2023neuroscienceinspiredscientificmachinepart2}:

\begin{equation}
L = \alpha L_2 + \gamma(S_M + S_P + S_{\text{spectral}} + S_{\text{spatial}} + S_f +S_\text{final})/6, 
\label{eq:loss}
\end{equation}

where $L_2$ is the standard reconstruction error and $S_i$ represents the average, over all unique spiking layers for component $i$, of the ratio of observed layer spike count to the total possible layer spikes for T steps. Ideally, higher $\gamma$ corresponds to lower $S_i$ and energy while resulting in worse $L_2$ error, meaning Equation \ref{eq:loss} allows control of the efficiency-reconstruction balance for VS-GNO.

\section{Results}

\subsection{2D Heat Exchanger and Training Details}

For evaluation, we observe the steady state flow analysis of a two dimensional cross section of a three dimensional heat exchanger with enhanced heat transfer due to its highly irregular geometry with dimpled surface and wavy-insert design \cite{AHMED2024104583}. The benchmark follows the framework in Equation \ref{eq:framework} with the following sparse-to-dense reconstruction: the field prediction of pressure and velocity components at 3,977 evaluation nodes from two scalar inlet values, temperature $T_{in}$ and axial velocity $v_{in}$, and a 100-point discretization of an axial wall heat flux profile. With a reconstruction ratio ($\frac{3977\cdot4}{102}$) of approximately $156:1$, this application provides an ideal complex multi-output physics problem with sparse-to-dense reconstruction and highly irregular geometry. Training and test data were generated by ANSYS Fluent software \cite{AnsysFluent2024} with a train-validation-test split of 988, 248, and 310 unique examples. VS-GNO was trained and evaluated in its full form and spectral-only form with $L = 10$ layers, an intermediate width of $100$, and $64$ spectral modes. The spectral only model simply removes the spatial block and collaboration mapping $f$ with its residual skip. The activation functions $\sigma$ utilized by the VSN are either GeLU if it existed before spiking integration or simply the identity function otherwise. Other training hyperparameters or model and loss term configurations are identical to the results presented by VIRSO \cite{howes2026graphneuraloperatoredge}. For both training and evaluation, we used the NVIDIA GH200, provided by the DeltaAI cluster from the National Center for Supercomputing Applications (NCSA) \cite{Delta}, to present preliminary results on the capabilities of VS-GNO with backpropagation through time (BPTT) and fast sigmoid surrogate gradient\cite{10242251} . The full and spectral-only models were initially evaluated on 1 STS and compared to the original ANN model for four loss function configurations: $\alpha=1$ and $\gamma=[0,0.05,0.1,0.5]$. We also present a spectral model evaluated on 10 STS with $\alpha=1$ and $\gamma=0.5$ in order to evaluate the integration of temporal dynamics.

\subsection{Model Performance}

In Table \ref{tab:spec}, for the four loss function configurations ($\gamma$) and the spectral-only VS-GNO at 1 STS, we present the average relative $L_2$ error over all channels (pressure and velocity) along with the spiking percentages defined in Equation \ref{eq:loss} for individual model components (Eq. \ref{eq:vsgno}). We demonstrate that for all configurations, the spectral VS-GNO is capable of minimal performance degradation with reconstruction $L_2$ error under $3\%$ for all configurations and the lowest error at $0.71\%$ at $\gamma=0$ which is negligibly higher than the VIRSO baseline performance, $\approx0.4\%$, without spiking. The spectral model also demonstrates significantly low spiking that is at most $\approx24\%$ for an individual component for all configurations (spectral block VSN) with an average spiking of $\approx7\%$ over all components for $\gamma=0.5$. These results also confirm the expected $\gamma$-spiking trend with higher $\gamma$ corresponding to lower spiking and higher $L_2$ error. In addition, the spectral VS-GNO at $\gamma=0.5$ and 10 STS resulted in only slightly higher $L_2$ and average spiking over all components ($L_2=3.29\%$ and $S_{\text{avg}}=7.6\%$), indicating that larger STS offers potentially little while introducing increased energy and latency.

\begin{table}[H]
  \caption{Spectral VS-GNO results for varying $\gamma$ with relative $L_2$ error averaged over channels and component spiking}
  \centering
  \label{tab:spec}
  \begin{tabular}{cccccc}
    \toprule
    $\gamma$ & $L_2$ &$S_M$&$S_P$&$S_{\text{spectral}}$&$S_{\text{final}}$\\
    \midrule
    $0$ & 0.71\%& 4.8\% & 15.2\%  & 23.8\% &16.7\% \\
    $0.05$ & 1.42\%& 2.3\% & 5.1\%  & 20.2\% &14.4\% \\
    $0.1$ & 2.78\%& 3.5\% & 3.9\%  & 19.5\% &13.0\% \\
    $0.5$ & 2.87\%& 0.8\% & 2.5\%  & 16.0\% &8.7\% \\
  \bottomrule
\end{tabular}
\end{table}

Table \ref{tab:full} demonstrates the spiking performance of the full VS-GNO with spatial and collaborative analysis. At $\gamma=0,0.05$, we see slightly higher mean $L_2$ error ($\approx2\%$ or lower) at higher spiking rates most likely due to the introduction of more sophisticated analysis which could result in more communication especially when there is no spiking regularization $\gamma=0$ and higher training difficulty. For example, we found the $\gamma=0.05$ configuration resulted in an average spiking rate of around $15\%$, similar to the spectral-only model at $\gamma=0$. Nevertheless, these first two configurations provide minimal degradation for the complete VS-GNO model with $L_2$ as low as $1.04\%$ ($\gamma=0$), similar to the VIRSO baseline of $0.4\%$, and significantly lower spiking (max $39.8\%$ and average 24.5\% for $\gamma=0$) for all components. Although, this specific use case does not demonstrate a sizable contribution to reconstruction from the spatial layer, analysis on the feasibility of spiking spatial analysis is required since further applications with stronger high-frequency modes and larger scale will require the spatial calibration \cite{howes2026graphneuraloperatoredge}. With $\gamma=0.1$, we found a more noticeable degradation in performance with mean $L_2$ error above $10\%$ and higher component spiking average ($\approx 19\%$) than $\gamma=0.05$, breaking the expected spiking-$\gamma$ trend and emphasizing the difficulty of spiking integration for the full VS-GNO architecture.

\begin{table}[H]
  \caption{Full VS-GNO results for varying $\gamma$ with relative $L_2$ error averaged over channels and component spiking} 
  \centering
  \label{tab:full}
  \begin{tabular}{cccccccc}
    \toprule
    $\gamma$ & $L_2$ &$S_M$&$S_P$&$S_f$&$S_{\text{spectral}}$&$S_{\text{spatial}}$&$S_{\text{final}}$\\
    \midrule
    $0$ & 1.04\%& 12.9\% & 12.4\%  & 16.2\% &39.8\% &35.2\% & 30.3\% \\
    $0.05$ & 2.15\%& 3.52\% & 8.73\% & 10.8\% &33.4\% &15.6\% & 19.7\%  \\
    $0.1$ & 10.60\%& 3.94\% & 23.1\%  & 27.6\% &27.0\% &15.1\% & 18.7\%  \\
    $0.5$ & 8.02\%& 0\% & 3.8\%  & 4.2\% &22.3\% &5.7\% & 10.1\% \\
  \bottomrule
\end{tabular}
\end{table}

Moreover, we found that the spiking percentage for spectral, spatial, and final $Q$ layers is significantly higher that the embedding and initial mappings $M$ and $P$, likely because the bulk of the learning occurs within the spectral-spatial transformations and the final layer must retain sufficient information to produce accurate outputs. In contrast, the embedding and lifting stages serve as initial feature projections and can most likely tolerate reduced information flow. Interestingly, for the full VS-GNO at $\gamma = 0.5$, the embedding mapping $M$ exhibits zero spiking, with outputs collapsing to the bias parameters of the final linear layer, suggesting that the model converged to a regime where minimizing latent representation and relying primarily on bias optimization was required to approximate the operator for the given configuration.

\section{Conclusions and Further Work}


In this paper, we present a Variable Spiking Graph Neural Operator to address the scalability and computational concerns of previous graph-based methodologies. The spectral-only version of VS-GNO resulted in 0.71\% $L_2$ error, compared to the 0.4\% VIRSO baseline at around 15\% average spiking for $\alpha=1, \gamma=0$, indicating VS-GNO's potential for a significant reduction in energy without severe performance degradation. Although the full version of VS-GNO at the same configuration had higher $L_2$ error, it still performed well at low $\gamma$ with significantly reduced spiking. As a result, with VS-GNO, we have positioned variable spiking and graph-based methodologies further towards successful sparse-to-dense reconstruction on highly irregular geometries with practical edge deployment.

Beyond architectural refinements, training methodology remains a central question and direction for further work. Surrogate gradient algorithms have strong performance but present with gradient mismatch potentially leading to the difficult training seen with the full VS-GNO. ANN-to-SNN conversion risks large STS that result in undesirable latency and increased energy consumption. Further investigation is required into alternative methods that can preserve low-latency and energy consumption as well as address or remove gradient bias and mismatch. Moreover, the practical realization of VS-GNO and other spiking operators with measurable energy conservation on physical neuromorphic hardware (Intel Loihi 2 and SpiNNaker2) has yet to occur and requires investigation into the process of such integration. Advancing both training methodologies and hardware integration will be essential to positioning operators, especially the VS-GNO, closer towards realistic applications, with sophisticated sparse-to-dense reconstruction on irregular geometries, while enabling practical edge deployment.

\section{Acknowledgments}
This work was made possible by support from the National Center for Supercomputing Applications (NCSA) and the U.S. Department of Energy Office of Nuclear Energy, specifically the University Nuclear Leadership Program's Graduate Fellowship.

\bibliographystyle{unsrt}  
\bibliography{references}
\end{document}